\documentclass[10pt,twocolumn,letterpaper]{article}

\usepackage{cvpr}
\usepackage{times}
\usepackage{epsfig}
\usepackage{graphicx}
\usepackage{amsmath}
\usepackage{amssymb}

\usepackage{pifont}
\usepackage{color}
\usepackage{booktabs}
\usepackage{multirow}
\usepackage{subcaption}
\usepackage{gensymb}
\usepackage{bm}
\usepackage{float}
\usepackage[table]{xcolor}
\graphicspath{ {./images/} }

\usepackage[pagebackref=true,breaklinks=true,letterpaper=true,colorlinks,bookmarks=false]{hyperref}
\usepackage{subcaption}
 \cvprfinalcopy 

\def\cvprPaperID{612} 

\ifcvprfinal\fi
\begin{document}

\title{LiDARsim:  Realistic LiDAR Simulation by Leveraging the Real World}

\author{
  Sivabalan Manivasagam$^{1,2}$\quad Shenlong Wang$^{1,2}$ \quad Kelvin Wong$^{1,2}$ \quad Wenyuan Zeng$^{1,2}$ \\  \quad Mikita Sazanovich$^{1}$ \quad Shuhan Tan$^{1}$\quad Bin Yang$^{1,2}$   \quad Wei-Chiu Ma$^{1,3}$ \quad Raquel Urtasun$^{1,2}$\\ 
 $^{1}$Uber Advanced Technologies Group \quad $^{2}$University of Toronto \\$^{3}$Massachusetts Institute of Techonology\\
 \small\texttt{\{manivasagam, slwang, kelvin.wong,  wenyuan, sazanovich, shuhan, byang10, weichiu, urtasun\}@uber.com}
}

\maketitle

\begin{abstract}

We tackle the problem of producing realistic simulations of LiDAR point clouds, the sensor of preference for most self-driving vehicles.
We argue that, by leveraging real data, we can simulate the complex world more realistically compared to employing virtual worlds built from CAD/procedural models. Towards this goal, we first build a large catalog of 3D static maps and 3D dynamic objects by driving around several cities with our self-driving fleet. We can then generate scenarios by selecting a scene from our catalog and "virtually" placing the self-driving vehicle (SDV) and a set of dynamic objects from the catalog in plausible locations in the scene. To produce realistic simulations, we develop a novel simulator that captures both the power of physics-based  and learning-based simulation. We first utilize ray casting over the 3D scene  and then use  a deep neural network to produce deviations from the physics-based simulation,  producing realistic LiDAR point clouds. We showcase LiDARsim's usefulness for perception algorithms-testing on long-tail events and end-to-end closed-loop evaluation on safety-critical scenarios. 

\end{abstract}

\vspace{-0.3cm}
\section{Introduction}

On a cold winter night, you are hosting a holiday gathering for friends and family. As the festivities come to a close, you notice that your best friend does not have a ride, so you request a self-driving car to take her home. 
You say farewell and go back inside to sleep, resting well knowing your friend is in good hands and will make it back safely.

\begin{figure}[t!]
\centering
\includegraphics[trim={0cm 0 0 0cm},clip, width=0.7\linewidth]{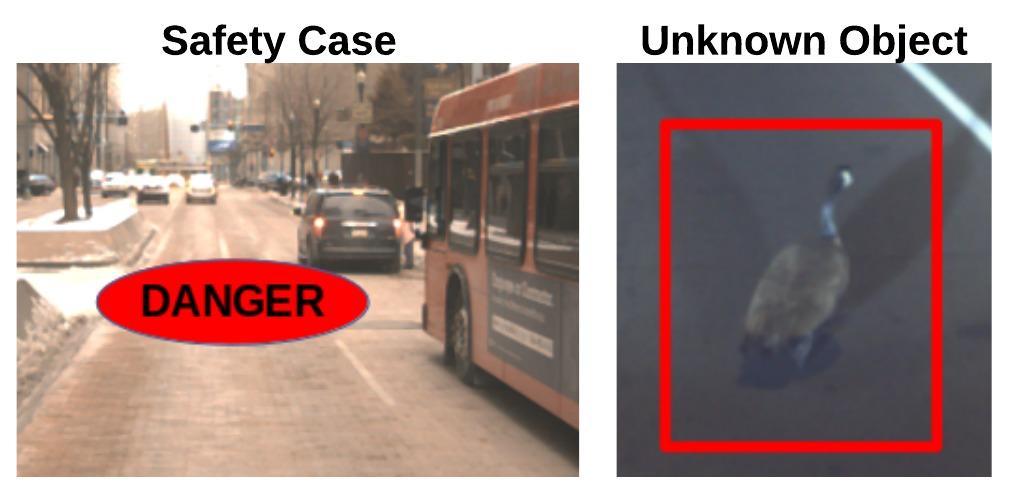}  
\caption{\textbf{Left}: What if a car hidden by the bus turns into our lane - can we avoid collision? \textbf{Right}: What if there is a goose on the road - can we detect it? See Fig. \ref{fig:safety_case_results} for results.}
\label{fig:safety_case}
\end{figure}

Many open questions remain to be answered 
to make self-driving vehicles (SDVs) a safe and trustworthy choice. How can we verify that the SDV can detect and handle properly objects it has never seen before (Fig.~\ref{fig:safety_case}, right)?  How do we guarantee that the SDV is robust and can maneuver safely in dangerous and safety-critical scenarios (Fig.~\ref{fig:safety_case}, left)? 

To make SDVs come closer to becoming a reality, 
we need to improve the safety of the autonomous system and demonstrate the safety case.
There are  three main approaches that the self-driving industry typically uses for improving and testing  safety: (1) real-world repeatable structured testing in a controlled environment, (2) evaluating on pre-recorded real-world data, (3) running experiments in simulation. 
Each of these approaches, while useful and effective, have limitations.

\begin{figure*}[t!]
\includegraphics[width=\linewidth]{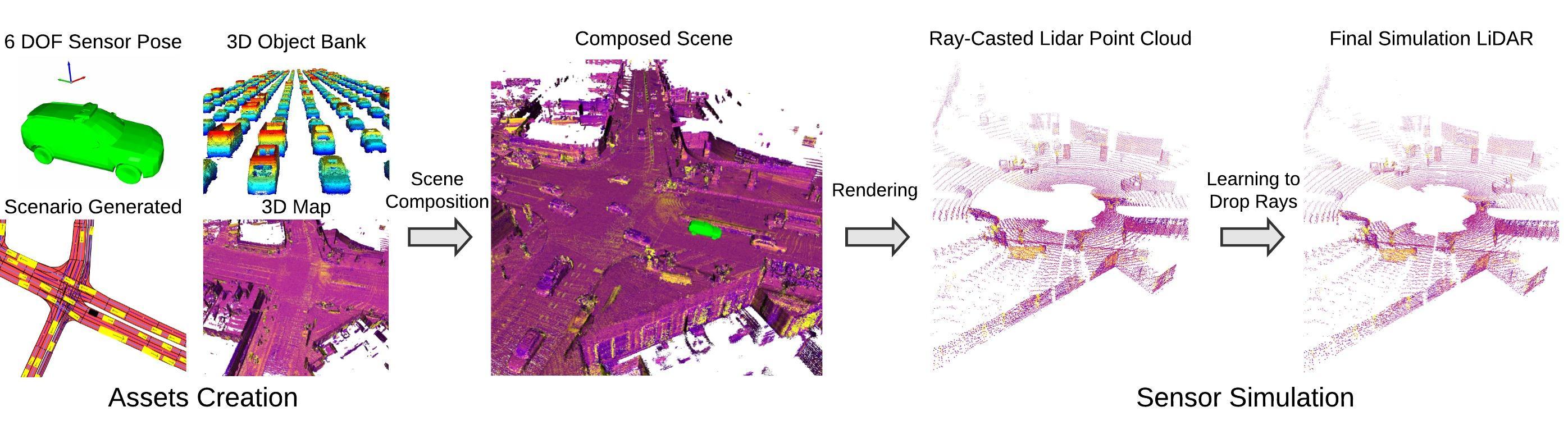}
   \caption{\textbf{LiDARsim Overview Architecture:} We first create the assets from real data, and then compose them into a scene and simulate the sensor with physics and machine learning.}
\label{fig:architecture}
\end{figure*}

Real-world testing in a structured environment, such as a test track, allows for full end-to-end testing of the autonomy system, but it is constrained to a very limited number of test cases, as it is very expensive and time consuming. 
Furthermore, safety-critical scenarios (i.e., mattress falling off of a truck at high speed, animals crossing street) are difficult to test safely and ethically.  
Evaluating on pre-recorded real-world data can leverage the high-diversity of real-world scenarios, but we can only collect data that we observe. 
Thus, the number of miles  necessary to collect sufficient long-tail events is way too large. 
Additionally, it is 
expensive to obtain labels. 
Like test-track evaluation, we can never fully test how the system will behave for scenarios it has never encountered and what the safety limit of the system is, which is crucial for demonstrating safety. 
Furthermore, because the data is prerecorded, this approach prevents the agent from interacting with the environment, as the sensor data will look different if the executed plan differs to what happened,  and thus it cannot be used to fully test the system performance. Simulation systems can in principle solve the limitations described above: closed-loop simulation can test how a robot would react under challenging and safety-critical situations, and we can use simulation to generate additional data for the long-tail events. 
Unfortunately, most existing simulation systems mainly focus on simulating behaviors and trajectories instead of simulating the sensory input, bypassing the perception module.  As a consequence, the full autonomy system cannot be tested, limiting the usefulness of these tests. 

However, if we 
could realistically simulate the sensory data, we could test the full autonomy system end-to-end.  We are not the first ones to realize the importance of sensor simulation; the history of simulating raw sensor data dates back to NASA and JPL's efforts supporting robot exploration of the surfaces of the moon and mars. 
Widely used robotic simulators, such as Gazebo and OpenRave \cite{koenig2004design, diankov2008openrave}, also support sensory simulation through physics and graphics engines. More recently, advanced real-time  rendering techniques have been exploited in autonomous driving simulators, such as CARLA and AirSim \cite{Dosovitskiy17, shah2018airsim}. However, their virtual worlds use handcrafted 3D assets and simplified physics assumptions resulting in simulations that do not represent well the statistics of real-world sensory data, resulting in a large sim-to-real domain gap. 

Closing the gap between simulation and the real-world  requires us to better model the real-world environment and the physics of the sensing processes. In this paper we focus on LiDAR, as it is the sensor of preference for most self-driving vehicles since it produces 3D point clouds from which 3D estimation is simpler and more accurate compared to using only cameras. Towards this goal, we propose LiDARsim, a novel, efficient, and realistic LiDAR simulation system. We argue that leveraging real data allows us to simulate LiDAR in a more realistic manner. LiDARsim has two stages: assets creation and sensor simulation (see Fig. \ref{fig:architecture}). At assets creation stage, we build a large catalog of 3D static maps and dynamic object meshes by driving around several cities with a vehicle fleet and accumulating information over time to get densified representations. This helps us simulate the complex world more realistically compared to employing virtual worlds designed by artists. At the sensor simulation stage, our approach combines the power of physics-based  and learning-based simulation. We first utilize raycasting over the 3D scene to acquire the initial physics rendering. Then, a deep neural network learns to deviate from the physics-based simulation to produce realistic LiDAR point clouds by learning to approximate more complex physics and sensor noise. 
 
The LiDARsim sensor simulator has a very small domain gap. 
This gives us the ability to test more confidently the full autonomy stack. 
We show in  experiments our perception algorithms' ability to detect unknown objects in the scene with LiDARsim. 
We also use LiDARsim to better understand how the autonomy system performs under safety-critical scenarios in a closed-loop setting that would be difficult to test without realistic sensor simulation. 
These experiments show the value that realistic sensory simulation can bring to self-driving. We believe this is just the beginning towards hassle-free testing and annotation-free training of self-driving autonomy systems.

\begin{figure*}
\includegraphics[width=\linewidth]{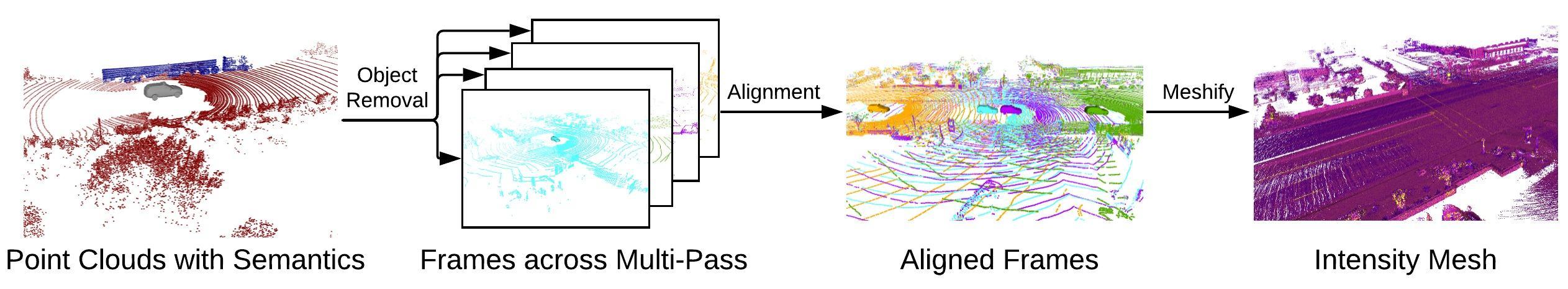}
   \caption{\textbf{Map Building Process:} We collect real data from multiple trajectories in the same area, remove moving objects, aggregate and align the data, and create a mesh surfel representation of the background.}
\label{fig:mapping}
\vspace{-0.0cm}
\end{figure*}


\section{Related Work}
\label{sec:related}

\paragraph{Virtual Environments:}
Virtual simulation environments are commonly used  in robotics and reinforcement learning. The seminal work of \cite{pomerleau1989alvinn}  trained a neural network on both real  and simulation data to learn to drive. Another popular direction is to exploit gaming environments, such as Atari games \cite{mnih2013playing}, Minecraft \cite{johnson2016malmo} and Doom \cite{kempka2016vizdoom}. However, due to unrealistic scenes and tasks that are evaluated in simple settings with few variations or noise, 
these types of environments do not generalize well to real-world. 3D virtual scenes have been used extensively for robotics in the context of navigation \cite{zhu2017target} and manipulation \cite{todorov2012mujoco, coumans2016pybullet}.
It is important for the agent trained in the simulation 
to generalize to the real world. Towards this goal, physics engines \cite{tan2018sim} have been exploited to mimic the real-world's physical interaction with the robot, such as multi-joint dynamics \cite{todorov2012mujoco} and vehicle dynamics \cite{wymann2000torcs}.   Another crucial component for virtual environment simulation is the quality of sensor simulation. The past decade has witnessed a significant improvement of real-time graphics engines such as Unreal \cite{games2007unreal} and Unity 3D \cite{engine2008unity}. 
Based on these graphics engines,  simulators have been developed to provide virtual sensor simulation such as CARLA and Blensor \cite{Dosovitskiy17, blensor, wang2019automatic, hurl2019precise, yue2018lidar}. However, there is still a large domain gap between the output of the simulators and the real world. 
We believe one reason for this domain gap is that the artist-generated environments are not diverse enough and the simplified physics models used 
do not account for important properties for sensor simulation such as material reflectivity or incidence angle of the sensor observation, which affect the output point cloud. For example, at most incidence angles, LiDAR rays will penetrate window glasses and not produce returns that can be detected by the receiver. 

\paragraph{Virtual Label Transfer:} Simulated data has great potential as it is possible to generate labels at scale mostly for free. 
This is appealing for tasks where labels are difficult to acquire such as optical flow and semantic segmentation  \cite{butler2012naturalistic, MIFDB16, ros2016synthia, richter2017playing,Gaidon:Virtual:CVPR2016}. Researchers have started to look into how to transfer an agent trained over simulated data to perform real-world tasks \cite{sun2018pwc,richter2017playing}.  It has been shown that pretraining over virtual labeled data can improve real-world perception performance, particularly 
when few or even zero real-world labels are available \cite{MIFDB16, richter2017playing, shrivastava2017learning, james2019sim2sim}.

\paragraph{Point Cloud Generation:}
Recent progress in generative models has provided the community with powerful tools for point cloud generation. \cite{ yang2019pointflow} transforms Gaussian-3D samples into a point cloud shape conditioned on class via normalizing flows, and \cite{caccia2019deep} uses VAEs and GANs to reconstruct LiDAR from noisy samples. In this work, instead of directly applying deep learning for point cloud generation or using solely graphics-based simulation, we adopt deep learning techniques to enhance graphics-generated LiDAR data, making it more realistic.  

\paragraph{Sensor Simulation in Real World:} While promising, 
past simulators have limited capability of mimicking the real-world, limiting their success to improve robots' real-world perception. This is because the virtual scene,  graphics engine, and physics engine are a simplification of the real-world.  

Motivated by this, recent work has started to bring real-world data into the simulator.  \cite{alhaija2018augmented} 
adds graphics-rendered dynamic objects to real camera images. Gibson Environment \cite{xia2018gibson, xia2019interactive} created an interactive simulator with rendered images that come from a RGBD scan of the real-world's indoor environment. 
Deep learning has been adopted to make the simulated images more realistic. Our work is related to Gibson environments, but our focus is on LiDAR sensor simulation over driving scenes. \cite{tallavajhula2018off} leveraged real data for creating assets of vegetation and road for off-road terrain lidar simulation. We would like to extend this to urban driving scenes and ensure realism for perception algorithms.
Very recently, in concurrent work, \cite{fang2018simulating} showcased  LiDAR simulation through raycasting over a 3D scene composed of 3D survey mapping data and CAD models. Our approach differs in several components: 1) We use a single standard LiDAR to build the map, as opposed to comprehensive 3D survey mapping, allowing us to map at scale 
in a cost effective manner (as our LiDAR is at least 10 times cheaper); 2) we build 3D objects from real-data, inducing more diversity and realism than CAD models (as shown in sec. \ref{sec:experiments}); 3) we utilize a learning system that models the residual physics not captured by graphics rendering to further boost the realism as opposed to standard rendering + random noise.


\section{Reconstructing the World for Simulation}

\begin{figure*}
\centering
  \includegraphics[width=\linewidth]{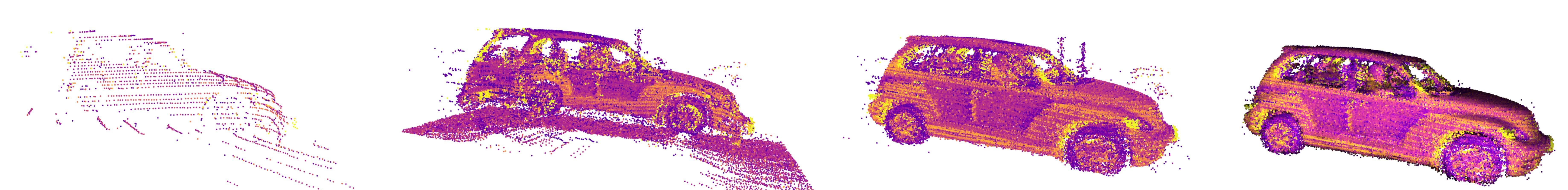}
  \caption{\textbf{Dynamic Object Creation:} From left to right: Individual sweep, Accumulated cloud, Symmetry completion, outlier removal and surfel meshing}
\label{fig:dyn}
\end{figure*}

Our objective is to build a LiDAR simulator that simulates complex scenes with many actors and produce point clouds with realistic geometry. 
We argue that by leveraging real data, we can simulate the world more realistically than when employing virtual worlds built solely from CAD/procedural models. 
To enable such a simulation, we need to first generate  a catalog of both static environments as well as dynamic objects. 
Towards this goal, we generate high definition 3D 
backgrounds and dynamic object meshes by driving around several cities with our self-driving fleet. 
We first describe how we generate 3D 
meshes of the static environment. We then describe how to build a library of dynamic objects. 
In sec. \ref{sec:simulation} we will address how to realistically simulate the LiDAR point cloud for the constructed scene.  

\subsection{3D Mapping for Simulation}
\label{sec:mapping}
To simulate real-world scenes, we first utilize sensor data scans to build our representation of the static 3D world. 
We want our representation to provide us high realism about the world and describe the physical properties about the material and geometry of the scene.
Towards this goal,  we collected data by driving over the same scene multiple times. On average, a static scene is created from 3 passes.  
Multiple LiDAR  sweeps are then associated to  a common coordinate system 
(the map frame) using offline Graph-SLAM \cite{thrun2006graph} with multi-sensor fusion leveraging wheel-odometry, IMU, LiDAR and GPS. 
This provides us centimeter accurate dense alignments of 
the LiDAR sweeps. 
We automatically remove moving objects (e.g., vehicles, cyclists, pedestrians)  
with LiDAR segmentation \cite{chrisvoxel}.

 \begin{figure*}
\includegraphics[width=0.4\linewidth]{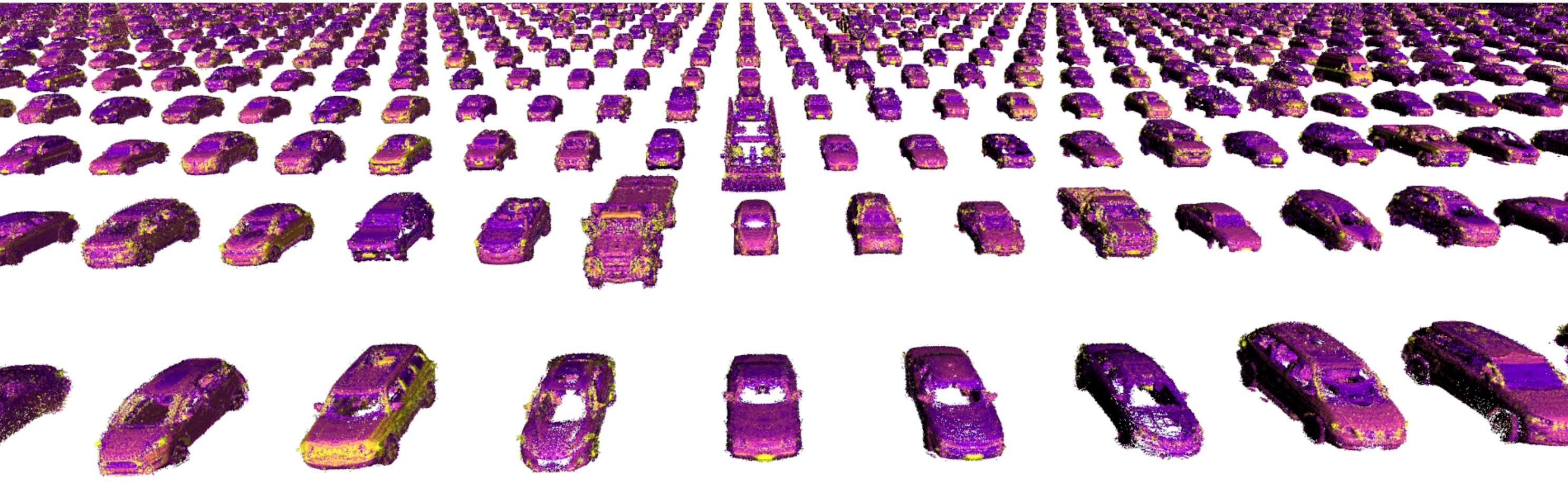}
\includegraphics[width=0.6\linewidth]{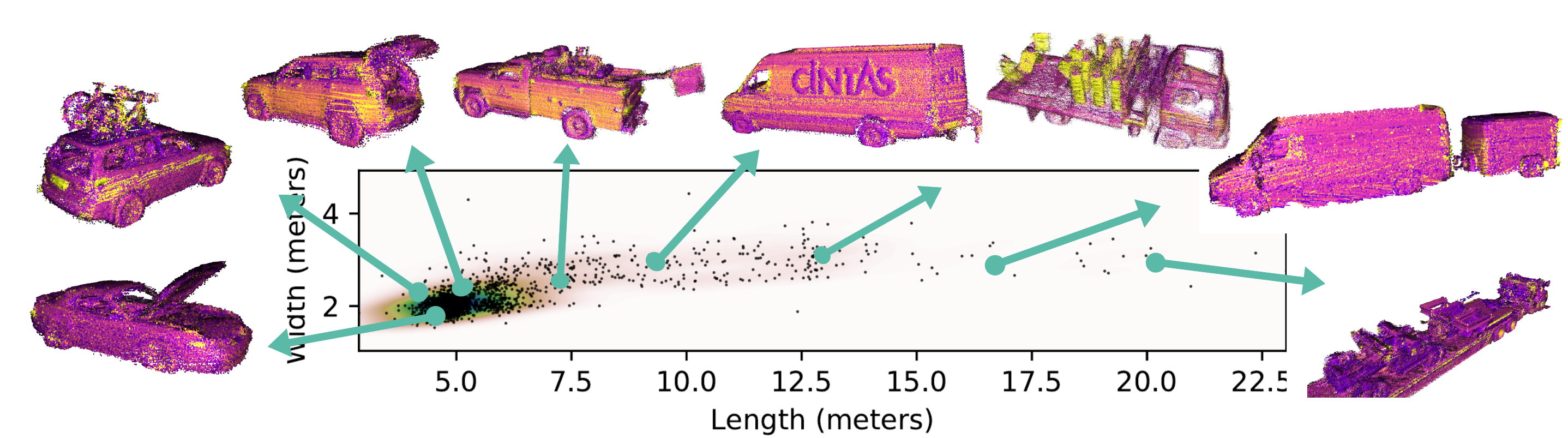}
  \caption{\textbf{Left}: Scale of our vehicle bank (displaying several hundred vehicles out of 25000), \textbf{Right}: Diversity of our vehicle bank colored by intensity, overlaid on vehicle dimension scatter plot; Examples (left to right): opened hood, bikes on top of vehicle, opened trunk, pickup with bucket, intensity shows text,  traffic cones on truck, van with trailer,  tractor on truck }
\label{fig:dyn_obj}
\end{figure*}

We then convert the aggregated LiDAR point cloud from multiple drives into a surfel-based 3D mesh of the scene through voxel-downsampling and normal estimation. 
We use surfels due to their simple construction, effective occlusion reasoning, and efficient collision checking\cite{pfister2000surfels}.  
In particular, we first downsample the point cloud, ensuring that over each $4\times4\times4$ cm$^3$ space only one point is sampled.

 For each such point, normal estimation is conducted through principal components analysis over neighboring points (20 cm radius and maximum of 200 neighbors). 

A disk surfel is then generated with the disk centre to be the input point and disk orientation to be its normal direction. 
In addition to geometric information, we record additional metadata about the surfel that we leverage later on for enhancing the realism of the simulated LiDAR point cloud. 
We record each surfel's (1) intensity value, (2) distance to the sensor, (3) and incidence angle (angle between the LiDAR sensor ray and the disk's surface normal). Fig.~\ref{fig:mapping} depicts our map building process, where  a reconstructed map colored by recorded intensity is shown in the last panel.  Note that this map-generation process is cheaper than using 3D artists, where the cost is thousands of dollars per city block.

\subsection{3D Reconstruction of Objects for Simulation}
\label{sec:dyn_obj}

To create realistic scenes, we also need to simulate dynamic objects, such as vehicles, cyclists, and pedestrians. 

Similar to our maps in Sec. \ref{sec:mapping}, we  leverage the real world to construct dynamic objects, where we can encode 
complicated physical phenomena not accounted for by raycasting via the recorded geometry and intensity metadata. 
We build a large-scale collection of dynamic objects using data collected from our self-driving fleet. We focus here on generating rigid objects such as vehicles, and in the future we will expand our method to deformable objects such as cyclists and pedestrians.
It is difficult to build full 3D mesh representations from sparse LiDAR scans due to the motion of objects and the partial observations captured by the LiDAR due to occlusion. 
We therefore develop a dynamic object generation process that leverages (1) inexpensive human-annotated labels, and (2) the symmetry of vehicles.

We exploit 3D bounding box annotations of objects over short 25 second snippets. Note that these annotations are prevalent in existing benchmarks such as KITTI or Nuscenes\cite{geiger2012we, caesar2019nuscenes}.

We  then accumulate the LiDAR points inside the bounding box and determine the object relative coordinates for the LiDAR points based on the bounding box center (see Fig.~\ref{fig:dyn}, second frame). This is not sufficient as this process often results in incomplete shapes due to partial observations. Motivated by the symmetry of vehicles, we mirror the point cloud along the vehicle's heading axis and concatenate with the raw point cloud. This gives a more complete shape as shown in Fig.~\ref{fig:dyn}, third frame. To further refine the shape and account for errors in point cloud alignment for moving objects, we apply an iterative color-ICP algorithm, where we use recorded intensity as the color feature \cite{park2017colored}. We then meshify the object through surfel-disk reconstruction, producing Fig.~\ref{fig:dyn}, last frame.
Similar to our approach with static scenes, we record intensity value, original range, and incidence angles of the surfels. 
Using this process, we generated a collection of over 25,000 dynamic objects. A few interesting objects are shown in Fig.~\ref{fig:dyn_obj}.  
We plan to release the generated assets to the community.

\section{Realistic Simulation for Self-driving}
\label{sec:simulation}
Given a traffic scenario, we compose the virtual world scene by placing the dynamic object meshes created in Sec.~\ref{sec:dyn_obj} over the 3D static environment from Sec.~\ref{sec:mapping}.

We now explain the physics-based simulation used to simulate both geometry and intensity of the LiDAR point cloud given sensor location, 3D assets, and traffic scenario as input. 
Then, we go over the features and data provided to a neural network to enhance the realism of the physics-based LiDAR point cloud by estimating which LiDAR rays do not return back to the sensor, which we call "raydrop".

\begin{figure*}[t!]
    \centering
\begin{tabular}{lcl}
       \includegraphics[trim={0cm 0 0 0cm},clip, width=0.48\linewidth]{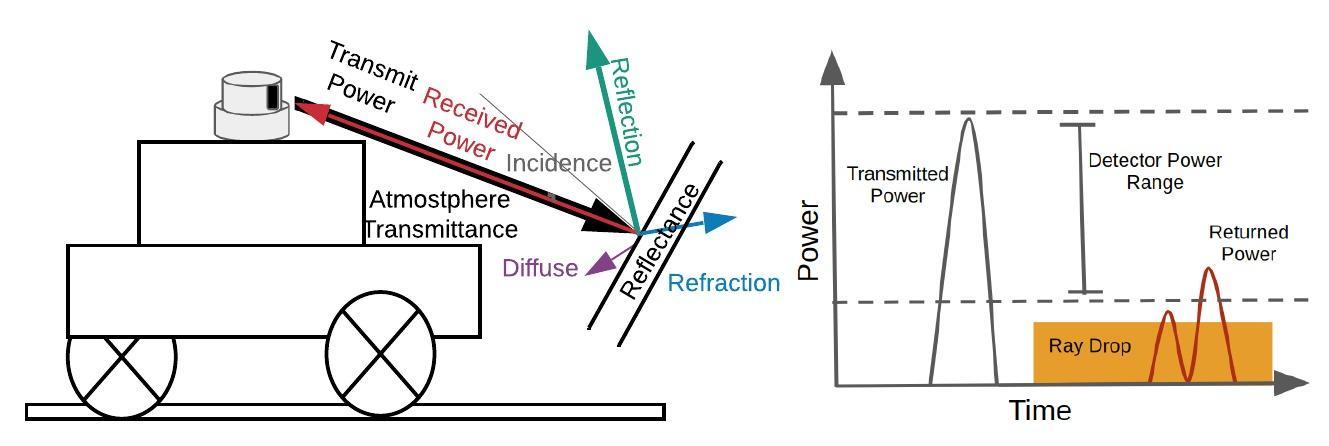}  
       \hspace{0.1cm}
       \includegraphics[width=0.48\linewidth]{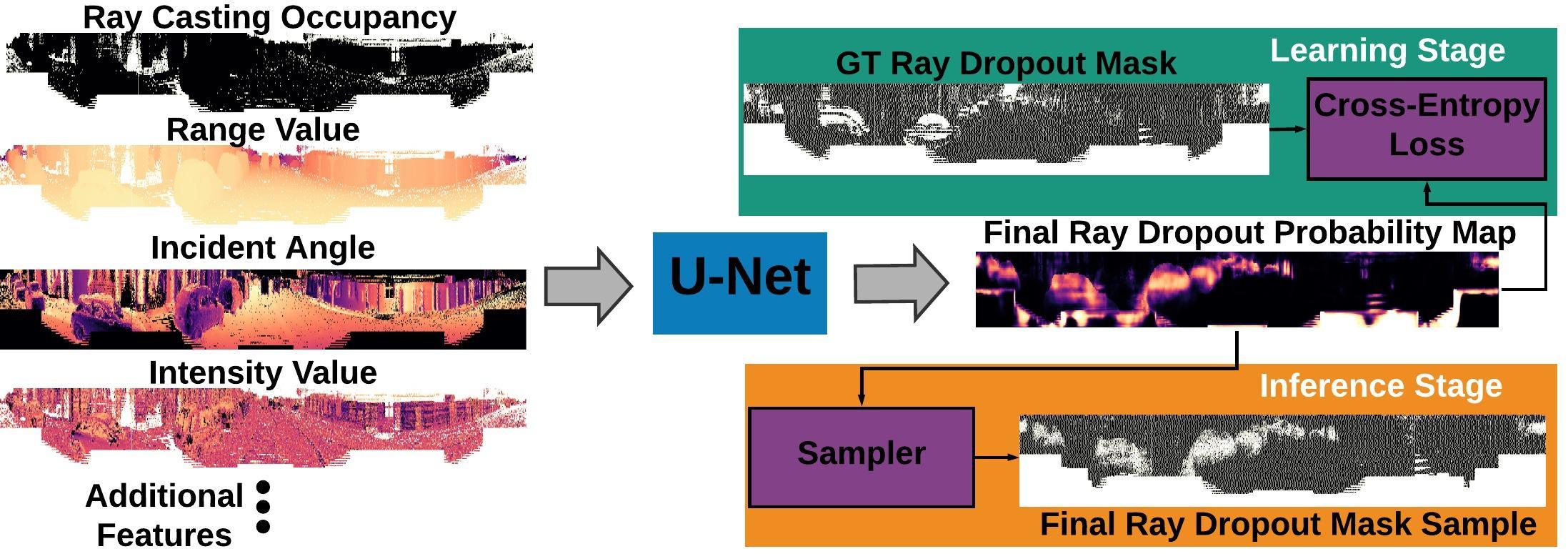}
\end{tabular}
    \caption{\textbf{Left}: Raydrop physics explained: Multiple real-world factors and sensor biases determine if the signal is detected by LiDAR receiver. \textbf{Right}: Raydrop network: Using ML and real data to approximate the raydropping process.}
\label{fig:raydrop}
\end{figure*}

\subsection{Physics based Simulation}
\label{sec:physics}
Our approach exploits physics-based simulation to create an estimation of the geometry of the generated point cloud. 
We focus on simulating a  scanning LiDAR, i.e., Velodyne HDL-64E, which is commonly used in many autonomous cars and  benchmarks such as KITTI \cite{geiger2012we}. The system has 64 emitter-detector pairs, each of which uses light pulses to measure distance. The basic concept is that each emitter emits a light pulse which travels until it hits a target, and a portion of the light energy is reflected back and received by the detector. Distance is measured by calculating the time of travel. 
The entire optical assembly rotates on a base to provide a 360-degree azimuth field of view at around 10 Hz with each full "sweep" providing approximately 110k returns. Note that none of the techniques described in this paper are restricted to this sensor type. 

We  simulate our LiDAR sensor with a graphics engine given a desired  6-DOF pose and velocity. Based on the LiDAR sensor's intrinsics parameters (see \cite{manual2014high} for sensor configuration), a set of rays are raycasted from the virtual LiDAR center into the scene. We simulate the rolling shutter effect by compensating for the ego-car's relative motion during the LiDAR sweep.  
Thus, for each ray shot from the LiDAR sensor at a vertical angle $\theta$ and horizontal angle $\phi$ we represent the ray with the source location $\mathbf{c}$ and shooting direction $\mathbf{n}$: 
$\mathbf{c} = \mathbf{c}_0 + (t_1 - t_0) \mathbf{v}_0$, $\mathbf{n} = \mathbf{R}_0[\cos\theta \cos\phi, \cos\theta \sin\phi, \sin \theta]^T$
where $\mathbf{c}_0$ is the sensor-laser's 3D location, $\mathbf{R}_0$ is the 3D rotation at the beginning of the sweep w.r.t. the map coordinates, $\mathbf{v}_0$ is the velocity and $t_1 - t_0$ is the change in time of the simulated LiDAR rays. In addition to rolling-shutter effects from the ego-car, we simulate the motion-blur of other vehicles moving in the scene during the LiDAR sweep. 
To balance computational cost with realism, we update objects poses within the LiDAR sweep at 360 equally spaced time intervals.
Using Intel Embree raycasting engine (which uses the Moller-Trumbore intersection algorithm \cite{moller2005fast}), we compute the ray-triangle collision against all surfels in the scene and find the closest surfel to the sensor that is hit.

Applying this to all rays in the LiDAR sweep, we obtain a physics-generated point cloud  
over the constructed scene. We also apply a
mask to remove rays that hit the SDV.  
\subsection{Learning to Simulate Raydrop}
\label{sec:learn}

\begin{figure}
\includegraphics[trim={0cm 0 0 0cm},clip, width=\linewidth]{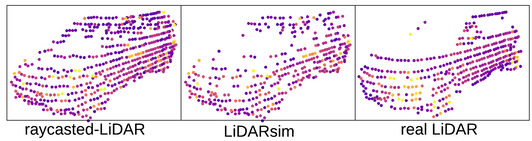}  
\caption{Qualitative Examples of Raydrop}
\label{fig:raydrop_qualitative}
\end{figure}

\paragraph{Motivation:} The LiDAR simulation approach described so far produces visually realistic geometry  
for LiDAR point clouds at first glance. However, we observe that the real LiDAR usually has approximately 10\% fewer LiDAR points that the raycasted version generated, and some vehicles have many more simulated LiDAR points than real. 
One assumption of the above physics-based approach is that every ray casted into the virtual world returns if it intersects with a physical surface. However, a ray casted by a real LiDAR sensor may not return (raydrop) if the strength of the return signal (the intensity value) is not strong enough to be detected by the receiver (see Fig. \ref{fig:raydrop}, left)\cite{kashani2015review}.
Modelling raydrop is a binary version of intensity simulation - it is a sophisticated and stochastic phenomenon impacted by factors such as material reflectance, incidence angle, range values, beam bias and other environment factors. Many of these factors are not available in artist-designed simulation environments, but leveraging real world data allows us to capture information, albeit noisy, about these factors. 
We frame LiDAR raydrop as a binary classification problem. We apply a neural network to learn the sensor's raydrop characteristics, utilizing machine learning to bridge the gap between simulated and real-world LiDAR data. 
Fig.~\ref{fig:raydrop}, right, summarizes the overall architecture.  
We next describe the model design and learning process.
\paragraph{Model and Learning:} 
To predict LiDAR raydrop, we transform the 3D LiDAR point cloud into a 64 x 2048 2D polar image grid, 
allowing us to encode which rays did not return from the LiDAR sensor, while also providing a mapping between the real LiDAR sweep and the simulated one (see Fig ~\ref{fig:raydrop}, right). 
We provide as input to the network a set of   channels\footnote{We use real-valued channels: range, \textit{original} recorded intensity, incidence angle, \textit{original} range of surfel hit, and \textit{original} incidence angle of surfel hit. Note that we obtained the \textit{original} values from the metadata recorded in sec. \ref{sec:mapping}, \ref{sec:dyn_obj}.  Integer-valued channels: laser id, semantic class (road, vehicle, background). Binary channels: Initial occupancy mask.}
 representing observable factors potentially influencing each ray's chance of not returning. 
 Our network architecture is a standard 8-layer U-Net \cite{ronneberger2015unet}. 
 The output of our network is a probability  for each element in the array if it returns or not. 
To simulate LiDAR noise, we sample from the probability mask to generate the output LiDAR point cloud.  We sample the probability mask instead of doing direct thresholding for two reasons: (1)  We learn raydrop with cross-entropy loss, meaning the estimated probabilities may not be well calibrated \cite{guo2017calibration} - sampling helps mitigate this issue compared to thresholding. (2) Real lidar data is non-deterministic due to additional noises (atmospheric transmittance, sensor bias) that our current approach may not fully model.  
As shown in Fig. \ref{fig:raydrop_qualitative},  learning raydrop  creates point clouds that better match the real data. 
\section{Experimental Evaluation}
\label{sec:experiments}
In this section we first introduce the city driving datasets that we apply our method on and cover LiDARsim implementation details. We then evaluate LiDARsim in four stages:  
(1) We demonstrate it is a high-fidelity simulator by comparing against the popular LiDAR simulation system CARLA via public evaluation on the KITTI dataset for segmentation and detection.
(2) We  evaluate LiDARsim against real LiDAR and simulation baselines on segmentation and vehicle detection. 
(3) We combine LiDARsim data with real data to further boost performance on perception tasks. 
(4) We showcase using LiDARsim to test  instance segmentation of unknown objects and end-to-end testing of the autonomy system in  safety critical scenarios.

\subsection{Experimental Setting}
\label{sec:exp_setting}
We evaluated our LiDAR simulation pipeline on a novel large-scale city dataset as well as  KITTI \cite{geiger2012we, behley2019semantickitti}. Our city dataset consists of 5,500 snippets of 25 seconds and ~1.4 million LiDAR sweeps captured at various seasons across the year. They contain multiple metropolitan cities in North America covering diverse scenes. 
Centimeter level localization is conducted through an offline process. 
We split our city dataset into 2 main sets: map-building ($\sim$87\%), and downstream perception (training $\sim$7\%, validation $\sim$1\%, and test $\sim$5\%). 
To accurately compare LiDARsim against real data, we simulate each real LiDAR sweep example using the SDV ground truth pose and the dynamic object poses based on the groundtruth scene layout for that sweep. Then for each  dynamic object we simulate, we compute a fitness score for each object in our library based on bounding box label dimensions and initial relative orientation to the SDV, and select a random object from the top scoring objects to simulate. 
 We then use the raycasted LiDAR sweep as input to train our raydrop network, and the respective real LiDAR sweep counterpart is used as labels. To train the raydrop network, we use 6 \% of snippets from map-building and use back-propagation with Adam \cite{adam} 
with a learning rate of $1e-4$.  
The view region for perception downstream tasks is -80 m. to 80m. along the vehicle heading direction and -40 m. to 40 m. orthogonal to heading direction. 

\begin{table}[]
\centering
\scalebox{.9}{
\begin{tabular}{lcccc}
\specialrule{.2em}{.1em}{.1em}
Train Set & Overall & Vehicle& Background \\ \hline
CARLA\textsuperscript{\cite{yoon2019mapless}} (Baseline) & 0.65 &  0.36  & 0.94  \\ 
LiDARsim (Ours) & 0.89 &  0.79  & 0.98 \\
SemanticKITTI (Oracle)  & 0.90 & 0.81   &  0.99\\
\specialrule{.2em}{.1em}{.1em}
\end{tabular}
}
\caption{LiDAR Vehicle Seg. (mIOU); SemanticKITTI val.}
\label{table:carla}
\end{table}

\begin{table}[]
\centering
\scalebox{.9}{
\begin{tabular}{lcccc}
\specialrule{.2em}{.1em}{.1em}
Train Set & IoU 0.5& IoU 0.7 \\ \hline
CARLA-Default (Baseline) & 20.0 &  11.5  \\ 
CARLA-Modified (Baseline) & 57.4 &  42.2  \\
LiDARsim (Ours) & 84.6 & 73.7  \\
KITTI (Oracle) & 88.1 & 80.0  \\
\specialrule{.2em}{.1em}{.1em}
\end{tabular}
}
\caption{LiDAR Vehicle Det (mAP); KITTI hard val. }
\label{table:kitti}
\end{table}

\begin{table}[]
\centering
\scalebox{0.9}{
\begin{tabular}{lcc}
\specialrule{.2em}{.1em}{.1em}
&  \multicolumn{2}{c}{IoU 0.7} \\
Train Set (100k)    & \multicolumn{1}{c}{$\geq$ 1 pt} &\multicolumn{1}{c}{$\geq$ 10 pt} \\ \hline
Real 	                                                              &  \textbf{75.2} &  \textbf{80.2} \\ 
GT raydrop                                    &  72.3  &  78.5 \\ \hline
ML raydrop                                &  \textit{71.6} &   \textit{78.6}\\
Random raydrop         & 69.4  & 77.5 \\
No raydrop                             & 69.2  & 77.4 \\
\specialrule{.2em}{.1em}{.1em}
\end{tabular}
}
\caption{Raydrop Analysis; Vehicle Det (mAP); Real Eval. }
\label{table:detection}
\end{table}  

\begin{table}[]
\centering
\scalebox{0.9}{
\begin{tabular}{lcc}
\specialrule{.2em}{.1em}{.1em}
&  \multicolumn{2}{c}{IoU 0.7} \\
Train Set (100k)    & \multicolumn{1}{c}{$\geq$ 1 pt} &\multicolumn{1}{c}{$\geq$ 10 pt} \\ \hline
Real	                                                              &  \textbf{75.2} &  \textbf{80.2} \\ 
Real-Data Objects (Ours)                              &  \textit{71.6} &   \textit{78.6}\\
CAD Objects                       & 65.9  & 74.3 \\
\specialrule{.2em}{.1em}{.1em}
\end{tabular}
}
\caption{CAD vs. Ours; Vehicle Det (mAP); Real Eval. }
\label{table:detection_CAD}
\end{table}  

\begin{table}[]
\centering
\scalebox{0.9}{
\begin{tabular}{lcccc}
\specialrule{.2em}{.1em}{.1em}
&\multicolumn{4}{c}{Segmentation (mIOU)} \\
Train Set &\multicolumn{1}{c}{Overall} & \multicolumn{1}{c}{Vehicle}& \multicolumn{1}{c}{Background} &\multicolumn{1}{c}{Road} \\ \hline
Real10k      & 90.2 &  87.0  & 92.8 &  90.8\\
Real100k      &  96.1 &   95.7  & 97.0 &  95.7 \\ \hline 
Sim100k      & 91.9 &   91.3  & 93.5 & 90.9  \\
Sim100k Real10k     &  94.6 &   93.9& 95.8 &   94.0\\
Sim100k Real100k      & 96.3 &   95.9 &  97.1 &   95.8 \\
\specialrule{.2em}{.1em}{.1em}
\end{tabular}
}
\caption{Data Augmentation; Segmentation; Real Eval.}
\label{table:augmentation_seg}
\end{table}

\begin{table}[]
\centering
\scalebox{0.9}{
\begin{tabular}{lcc}
\specialrule{.2em}{.1em}{.1em}
& \multicolumn{2}{c}{IoU 0.7} \\
Train Set  & \multicolumn{1}{c}{ $\geq$ 1 pt} &  \multicolumn{1}{c}{$\geq$ 10 pt}   \\ \hline
Real 10k                                      & 60.0    &   65.9 \\
Real 100k                                    &75.2     &   80.2\\ \hline
Sim 100k                                     &71.1   & 78.1\\ 
Real 10k + Sim100k                  &73.5  & 79.8\\
Real 100k + Sim 100k               &77.6  &  82.2\\
\specialrule{.2em}{.1em}{.1em}
\end{tabular}
}
\caption{Data Augmentation; Vehicle Detection; Real Eval.}
\label{table:augmentation_det}
\end{table}

\subsection{Comparison against Existing Simulation}
\label{sec:exp_domain}
To demonstrate the realism of LiDARsim, 
we apply LiDARsim to the public KITTI benchmark for vehicle segmentation and detection and compare against the existing simulation system CARLA. 
We train perception models with simulation data and evaluate on KITTI.
To compensate for the domain gap due to labeling policy and sensor configurations between KITTI and our dataset,
we make the following modifications to LiDARsim: 
(1) adjust sensor height to be at KITTI vehicle height, 
(2) adjust azimuth resolution to match KITTI data, and 
(3) utilize KITTI labeled data to generate a KITTI dynamic object bank. 
Adjustments (1) and (2) are also applied to adapt CARLA under the KITTI setting (\textbf{CARLA-Default}). 
The original CARLA LiDAR simulation uses the collision hull to render dynamic objects, resulting in simplistic and unrealistic LiDAR. To improve CARLA's realism, we generate LiDAR data by sampling from the depth-image according to 
the Velodyne HDL-64E  setting (\textbf{CARLA-Modified}). The depth-image uses the 3D CAD model geometry, generating 
more realistic 
LiDAR.

\begin{figure}[t]
\centering
\includegraphics[trim={0 0 0 0},clip, width=1.0\linewidth]{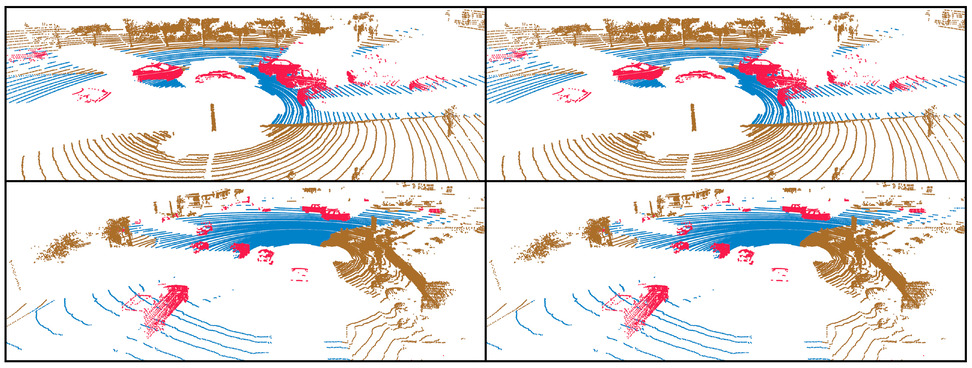}
   \caption{
Segmentation Segmentation on Real LiDAR point clouds. \textbf{Left}: LiDARsim trained; \textbf{Right}: real trained.  {\color{blue} Road}, {\color{red}Car}, {\color{brown}Background}} 
\label{fig:qualitative_seg}
\end{figure}

Table \ref{table:carla} shows vehicle and background segmentation evaluation on the SemanticKITTI dataset \cite{behley2019semantickitti} using the LiDAR segmentation network from  \cite{chrisvoxel}.  We train on 5k examples using either CARLA motion-distorted LiDAR \cite{yoon2019mapless}, LiDARsim using scene layouts from our dataset, or SemanticKITTI LiDAR, the oracle for our task. LiDARsim is very close to SemanticKITTI performance and significantly outperforms CARLA 5k. 
We also evaluate the performance on the birds-eye-view (BEV) vehicle detection task. Specifically, we simulate 100k frames of LiDAR training data using either LiDARsim or CARLA,  train a BEV detector \cite{yang2018hdnet}, and evaluate over KITTI validation set. For KITTI Real data, we use standard train/val splits and data augmentation techniques \cite{yang2018hdnet}.
As shown in Table \ref{table:kitti}  (evaluated at "hard" setting), LiDARsim outperforms CARLA and has close performance with the real KITTI data, despite being from different geographic domains.

\subsection{Ablation Studies}
\label{sec:exp_ablation}
We conduct two ablation studies to evaluate the use of real-world assets and the raydrop network. We train on either simulated or real data and then evaluate mean average precision (mAP) at IoU 0.7  at different LiDAR points-on-vehicle thresholds (fewer points is harder). 

\paragraph{Raydrop:}
We compare the use of our proposed raydrop network against three baselines:
"\textbf{No raydrop}," is raycasting with no raydrop; all rays casted to the scene that return are included in the point cloud. "\textbf{GT raydrop}," raycasts only the rays returned from the real LiDAR sweep. This serves as an oracle performance of our ray drop method.
"\textbf{Random raydrop}," randomly drops 10\% of the raycasted LiDAR points, as this is the average difference in returned points between real LiDAR and \textbf{No raydrop} LiDAR. 
As shown in Tab.~\ref{table:detection} using \textbf{"ML Raydrop"} boosts detection by 2\% AP compared to raycasting or random raydrop, and is close to oracle  \textbf{"GT Raydrop"} performance. 

\begin{figure}[t]
\centering
\includegraphics[width=1.0\linewidth]{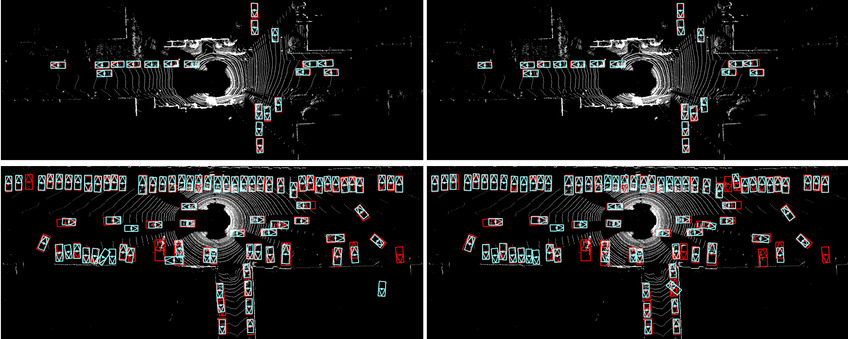}
   \caption{BEV Detection on real LiDAR point clouds. \textbf{Left:} LiDARsim trained; \textbf{Right}: real trained. {\color{cyan} Blue: Predictions}, {\color{red}Red: Groundtruth}}
\label{fig:qualitative_detection}
\end{figure}

\paragraph{Real Assets vs CAD models:}
Along with evaluating different data generation baselines, we also evaluate the use of real data to generate dynamic objects. Using the same LiDARsim pipeline, we replace our dynamic object bank with a bank of 140 vehicle CAD models. Bounding box labels for the CAD models are generated by using the same bounding box as the closest object in our bank based on point cloud dimensions.  As shown in Tab.~\ref{table:detection_CAD}, LiDARsim with CAD models has a larger gap (9\% mAP gap) with real data vs. LiDARsim with real-data based objects (3.6\% gap).

\subsection{Combining Real and LiDARsim Data}
\label{sec:exp_augmentation}
We now combine real data with LiDARsim data generated from groundtruth scenes to see if simulated data can further boost performance when used for training. 
As shown in Tab.~\ref{table:augmentation_seg},  with a small number of real training examples, the network's performance degrades. However, with the help of simulated data, even with around $10\%$ real data, we are able to achieve similar performance as 100\% real data, with less than $1\%$ mIOU difference, highlighting LiDARsim's potential to reduce the cost of annotation. When we have large-scale training data, simulation data offers marginal performance gain for vehicle segmentation. 
Tab.~\ref{table:augmentation_det} shows the mAP of object detection  using simulated training data. Compared against using 100k training data, augmenting with simulated data helps further boost the performance. 

\subsection{LiDARsim for Safety and Edge-Case Testing}
\label{sec:exp_safety}
We conduct three experiments to demonstrate LiDARsim for edge-case testing and safety evaluation. We first evaluate LiDARsim's coherency against real data when it is used as testing protocol for models trained only with real data. We then test perception algorithms on LiDARsim for identifying unseen rare objects. Finally, we demonstrate how LiDARsim allows us to evaluate how a motion planner maneuvers safety-critical scenarios in a closed-loop setting.

\begin{table}[t]
\centering
\begin{tabular}{lcc}
\specialrule{.2em}{.1em}{.1em}
Metric   & \multicolumn{1}{c}{IoU 0.5} & \multicolumn{1}{c}{IoU 0.7} \\ \hline
Eval on Real (AP)	                                                      &  91.5   &  75.2 \\
Eval on LiDARsim (AP) &  90.2   &  77.9   \\ \hline
Detection Agreement &  94.7   & 86.5  \\
\specialrule{.2em}{.1em}{.1em}
\end{tabular}
\caption{Performance gap between evaluating on sim. data vs. real data for model trained only on real data ($\geq$ 1 pt)}
\label{table:detection_sim}
\end{table}
\paragraph{Real2Sim Evaluation:}
 To demonstrate that LiDARsim could be used to directly evaluate a model trained solely on real data, we report in Tab. \ref{table:detection_sim} results for a detection model trained on 100k real data and evaluated on either the Real or LiDARsim test set. We also report a new metric called ground-truth detection agreement: 
$\kappa_{det} = \frac{\mid R_+ \cap S_+ \mid + \mid R_- \cap S_- \mid}{\mid R_+ \cup R_- \mid}$, 
where $R_+$ and $R_-$ are the sets of  ground-truth labels that are detected and missed, respectively, when the model is evaluated on real data, and $S_{+}$ and $S_-$,  when the model is evaluated on simulated data. With a paired set of ground-truth labels and detections,  we ideally want $\kappa_{det} = 1$, where a model evaluated on either simulated or real data produces the same set of detections \textit{and} missed detections. At IoU=0.5, almost 95\% of true detections and missed detections match in real and LiDARsim data. 

\paragraph{Rare Object Testing:}
 We now use LiDARsim to analyze 
a perception algorithm for the task of open-set panoptic segmentation: identifying known and unknown instances in the scene, along with semantic classes that do not have instances, such as background or road. 
We evaluate OSIS \cite{wong2019identifying} to detect unknown objects. We utilize CAD models of animals and construction elements that we place in the scene to generate 20k unknown-object evaluation LiDAR sweeps. We note that we use CAD models here since we would like to evaluate OSIS's ability to detect unknown objects that the vehicle has never observed.

\begin{figure}[t]
\centering
\includegraphics[trim={0cm 0 0 0cm},clip, width=\linewidth]{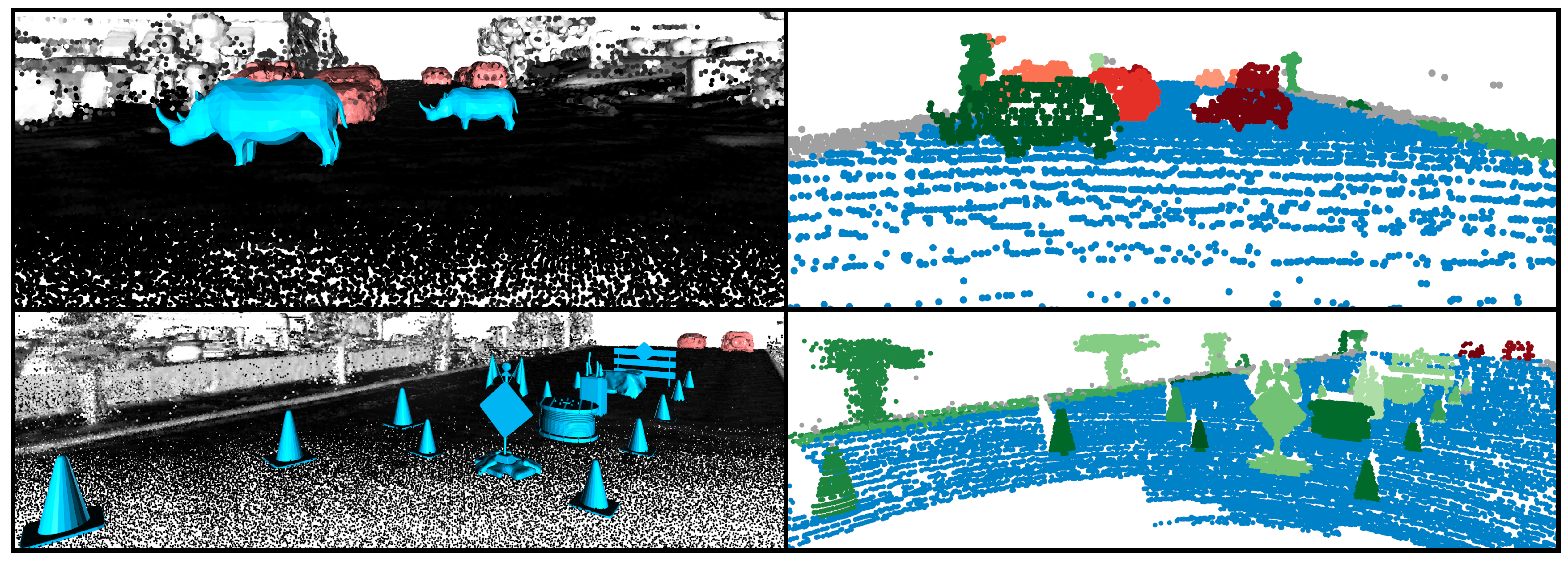}
\caption{Evaluating perception for unknown objects.  Trained on real, evaluated with LiDARsim.  \textbf{Left}: Simulated scene. \textbf{Right}: OSIS Segmentation Predictions. Unkown instances are in shades of green. A rhino is incorrectly detected as car (red). Construction correctly detected. }
\label{fig:open_set_results}
\end{figure}

\begin{figure}[]
\centering
\includegraphics[trim={0cm 0 0 0cm},clip, width=1.0\linewidth]{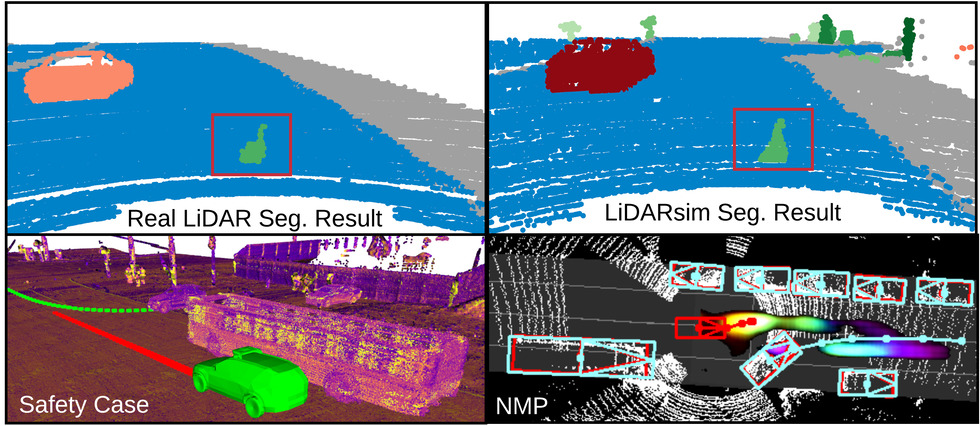}
\caption{Results for cases in Figure \ref{fig:safety_case}.  Models trained on real, evaluated on LiDARsim. \textbf{Top-left}: OSIS on Real  \textbf{Top-right}: OSIS on LiDARsim \textbf{Bot-left}: Safety Case in LiDARsim \textbf{Bot-right}: NMP Planned path to avoid collision}
\label{fig:safety_case_results}
\end{figure}

 We leverage the lane graph of our scenes to create different types of scenarios: animals crossing a road, construction blocking a lane, and random objects scattered on the street. Table \ref{table:open_sim} shows reported unknown and panoptic quality (UQ/PQ) for an OSIS model trained only on real data. 
Our qualitative example in Fig. \ref{fig:safety_case_results} shows OSIS's performance on real and LiDARsim closely match: OSIS detects the goose.  We are also able to identify situations where OSIS 
can improve, such as in Fig. \ref{fig:open_set_results}: a crossing rhino is incorrectly segmented as a vehicle.

\paragraph{Safety-critical  Testing:}
We  now evaluate perception on the end-metric performance of the autonomy system: safety. We evaluate an enhanced neural motion planner's (NMP)  \cite{zeng2019end} ability to maneuver safety-critical scenarios. 

We take the safety-critical test case described in Fig. \ref{fig:safety_case_results} and generate 110 scenarios of the test case in geographic areas in different cities and traffic configurations. To understand the safety buffers of NMP, we vary the initial velocity of the SDV and the trigger time of the occluded vehicle entering the SDV's lane. 
Fig. \ref{fig:safety_case_results} shows qualitative results. 
On average, the research prototype NMP succeeds 90\% of the time.

\begin{table}[]
\centering
\begin{tabular}{lccc}
\specialrule{.2em}{.1em}{.1em}
&  \multicolumn{1}{c}{Unknown (UQ)}& \multicolumn{1}{c}{Vehicle (PQ)}& \multicolumn{1}{c}{Road (PQ)}\\ \hline
LiDARsim &  54.9 & 87.7 & 93.4\\
Real\textsuperscript{\cite{wong2019identifying}} & 66.0 & 93.5 & 97.7\\
\specialrule{.2em}{.1em}{.1em}
\end{tabular}
\caption{Open-set Seg. results, trained on real LiDAR}
\label{table:open_sim}
\end{table}

\vspace{-0.2cm}
\section{Conclusion}

LiDARsim leverages real-world data, physics, and machine learning to simulate realistic LiDAR sensor data. 
With no additional training or domain adaptation, we 
can
directly apply perception algorithms trained on real data and evaluate them with LiDARsim in novel and safety-critical scenarios, achieving results that match closely with the real world and gaining new insights into the autonomy system.
Along with enhancing LiDARsim with intensity simulation and conditional generative modeling for weather conditions, we envision using LiDARsim for  end-to-end training and testing in simulation, opening the door to reinforcement learning and imitation learning for self-driving.
We plan to share LiDARsim with the community to help develop more robust and safer solutions to self-driving.

{\small
\bibliographystyle{ieee}
\bibliography{egbib_new}

\begin{thebibliography}{10}\itemsep=-1pt

\bibitem{alhaija2018augmented}
H.~A. Alhaija, S.~K. Mustikovela, L.~Mescheder, A.~Geiger, and C.~Rother.
\newblock Augmented reality meets computer vision: Efficient data generation
  for urban driving scenes.
\newblock {\em IJCV}, 2018.

\bibitem{behley2019semantickitti}
J.~Behley, M.~Garbade, A.~Milioto, J.~Quenzel, S.~Behnke, C.~Stachniss, and
  J.~Gall.
\newblock Semantickitti: A dataset for semantic scene understanding of lidar
  sequences.
\newblock In {\em ICCV}, 2019.

\bibitem{butler2012naturalistic}
D.~J. Butler, J.~Wulff, G.~B. Stanley, and M.~J. Black.
\newblock A naturalistic open source movie for optical flow evaluation.
\newblock In {\em ECCV}, 2012.

\bibitem{caccia2019deep}
L.~Caccia, H.~van Hoof, A.~Courville, and J.~Pineau.
\newblock Deep generative modeling of lidar data.
\newblock 2019.

\bibitem{caesar2019nuscenes}
H.~Caesar, V.~Bankiti, A.~H. Lang, S.~Vora, V.~E. Liong, Q.~Xu, A.~Krishnan,
  Y.~Pan, G.~Baldan, and O.~Beijbom.
\newblock nuscenes: A multimodal dataset for autonomous driving.
\newblock {\em arXiv}, 2019.

\bibitem{coumans2016pybullet}
E.~Coumans and Y.~Bai.
\newblock Pybullet, a python module for physics simulation for games, robotics
  and machine learning.
\newblock {\em GitHub repository}, 2016.

\bibitem{diankov2008openrave}
R.~Diankov and J.~Kuffner.
\newblock Openrave: A planning architecture for autonomous robotics.
\newblock {\em Robotics Institute, Pittsburgh, PA, Tech. Rep. CMU-RI-TR-08-34},
  2008.

\bibitem{Dosovitskiy17}
A.~Dosovitskiy, G.~Ros, F.~Codevilla, A.~Lopez, and V.~Koltun.
\newblock {CARLA}: {An} open urban driving simulator.
\newblock In {\em CoRL}, 2017.

\bibitem{engine2008unity}
U.~G. Engine.
\newblock Unity game engine-official site.
\newblock {\em Online: http://unity3d. com}, 2008.

\bibitem{fang2018simulating}
J.~Fang, F.~Yan, T.~Zhao, F.~Zhang, D.~Zhou, R.~Yang, Y.~Ma, and L.~Wang.
\newblock Simulating lidar point cloud for autonomous driving using real-world
  scenes and traffic flows.
\newblock {\em arXiv}, 2018.

\bibitem{Gaidon:Virtual:CVPR2016}
A.~Gaidon, Q.~Wang, Y.~Cabon, and E.~Vig.
\newblock Virtual worlds as proxy for multi-object tracking analysis.
\newblock In {\em CVPR}, 2016.

\bibitem{games2007unreal}
E.~Games.
\newblock Unreal engine.
\newblock {\em Online: https://www. unrealengine. com}, 2007.

\bibitem{geiger2012we}
A.~Geiger, P.~Lenz, and R.~Urtasun.
\newblock Are we ready for autonomous driving? the kitti vision benchmark
  suite.
\newblock In {\em CVPR}, 2012.

\bibitem{blensor}
M.~Gschwandtner, R.~Kwitt, A.~Uhl, and W.~Pree.
\newblock Blensor: Blender sensor simulation toolbox.
\newblock In {\em ISVC}, 2011.

\bibitem{guo2017calibration}
C.~Guo, G.~Pleiss, Y.~Sun, and K.~Q. Weinberger.
\newblock On calibration of modern neural networks.
\newblock In {\em ICML}, 2017.

\bibitem{hurl2019precise}
B.~Hurl, K.~Czarnecki, and S.~Waslander.
\newblock Precise synthetic image and lidar (presil) dataset for autonomous
  vehicle perception.
\newblock {\em arXiv}, 2019.

\bibitem{james2019sim2sim}
S.~James, P.~Wohlhart, M.~Kalakrishnan, D.~Kalashnikov, A.~Irpan, J.~Ibarz,
  S.~Levine, R.~Hadsell, and K.~Bousmalis.
\newblock Sim-to-real via sim-to-sim: Data-efficient robotic grasping via
  randomized-to-canonical adaptation networks.
\newblock In {\em CVPR}, 2019.

\bibitem{johnson2016malmo}
M.~Johnson, K.~Hofmann, T.~Hutton, and D.~Bignell.
\newblock The malmo platform for artificial intelligence experimentation.
\newblock In {\em IJCAI}, 2016.

\bibitem{kashani2015review}
A.~G. Kashani, M.~J. Olsen, C.~E. Parrish, and N.~Wilson.
\newblock A review of lidar radiometric processing: From ad hoc intensity
  correction to rigorous radiometric calibration.
\newblock {\em Sensors}, 2015.

\bibitem{kempka2016vizdoom}
M.~Kempka, M.~Wydmuch, G.~Runc, J.~Toczek, and W.~Ja{\'s}kowski.
\newblock Vizdoom: A doom-based ai research platform for visual reinforcement
  learning.
\newblock In {\em CIG}, 2016.

\bibitem{adam}
D.~P. Kingma and J.~Ba.
\newblock Adam: A method for stochastic optimization.
\newblock {\em arXiv}, 2014.

\bibitem{koenig2004design}
N.~Koenig and A.~Howard.
\newblock Design and use paradigms for gazebo, an open-source multi-robot
  simulator.
\newblock In {\em IROS}, 2004.

\bibitem{manual2014high}
V.~Manual.
\newblock High definition lidar-hdl 64e user manual, 2014.

\bibitem{MIFDB16}
N.~Mayer, E.~Ilg, P.~H{\"a}usser, P.~Fischer, D.~Cremers, A.~Dosovitskiy, and
  T.~Brox.
\newblock A large dataset to train convolutional networks for disparity,
  optical flow, and scene flow estimation.
\newblock In {\em CVPR}, 2016.

\bibitem{mnih2013playing}
V.~Mnih, K.~Kavukcuoglu, D.~Silver, A.~Graves, I.~Antonoglou, D.~Wierstra, and
  M.~Riedmiller.
\newblock Playing atari with deep reinforcement learning.
\newblock {\em arXiv}, 2013.

\bibitem{moller2005fast}
T.~M{\"o}ller and B.~Trumbore.
\newblock Fast, minimum storage ray/triangle intersection.
\newblock In {\em ACM SIGGRAPH 2005 Courses}, 2005.

\bibitem{park2017colored}
J.~Park, Q.-Y. Zhou, and V.~Koltun.
\newblock Colored point cloud registration revisited.
\newblock In {\em ICCV}, 2017.

\bibitem{pfister2000surfels}
H.~Pfister, M.~Zwicker, J.~Van~Baar, and M.~Gross.
\newblock Surfels: Surface elements as rendering primitives.
\newblock In {\em SIGGRAPH}, 2000.

\bibitem{pomerleau1989alvinn}
D.~A. Pomerleau.
\newblock Alvinn: An autonomous land vehicle in a neural network.
\newblock In {\em NIPS}, 1989.

\bibitem{richter2017playing}
S.~R. Richter, Z.~Hayder, and V.~Koltun.
\newblock Playing for benchmarks.
\newblock In {\em ICCV}, 2017.

\bibitem{ronneberger2015unet}
O.~Ronneberger, P.~Fischer, and T.~Brox.
\newblock U-net: Convolutional networks for biomedical image segmentation.
\newblock In {\em MICCAI}, 2015.

\bibitem{ros2016synthia}
G.~Ros, L.~Sellart, J.~Materzynska, D.~Vazquez, and A.~M. Lopez.
\newblock The synthia dataset: A large collection of synthetic images for
  semantic segmentation of urban scenes.
\newblock In {\em CVPR}, 2016.

\bibitem{shah2018airsim}
S.~Shah, D.~Dey, C.~Lovett, and A.~Kapoor.
\newblock Airsim: High-fidelity visual and physical simulation for autonomous
  vehicles.
\newblock In {\em Field and service robotics}, 2018.

\bibitem{shrivastava2017learning}
A.~Shrivastava, T.~Pfister, O.~Tuzel, J.~Susskind, W.~Wang, and R.~Webb.
\newblock Learning from simulated and unsupervised images through adversarial
  training.
\newblock In {\em CVPR}, 2017.

\bibitem{sun2018pwc}
D.~Sun, X.~Yang, M.-Y. Liu, and J.~Kautz.
\newblock Pwc-net: Cnns for optical flow using pyramid, warping, and cost
  volume.
\newblock In {\em CVPR}, 2018.

\bibitem{tallavajhula2018off}
A.~Tallavajhula, {\c{C}}.~Meri{\c{c}}li, and A.~Kelly.
\newblock Off-road lidar simulation with data-driven terrain primitives.
\newblock In {\em ICRA}, 2018.

\bibitem{tan2018sim}
J.~Tan, T.~Zhang, E.~Coumans, A.~Iscen, Y.~Bai, D.~Hafner, S.~Bohez, and
  V.~Vanhoucke.
\newblock Sim-to-real: Learning agile locomotion for quadruped robots.
\newblock {\em arXiv}, 2018.

\bibitem{thrun2006graph}
S.~Thrun and M.~Montemerlo.
\newblock The graph slam algorithm with applications to large-scale mapping of
  urban structures.
\newblock {\em The International Journal of Robotics Research}, 2006.

\bibitem{todorov2012mujoco}
E.~Todorov, T.~Erez, and Y.~Tassa.
\newblock Mujoco: A physics engine for model-based control.
\newblock In {\em IROS}, 2012.

\bibitem{wang2019automatic}
F.~Wang, Y.~Zhuang, H.~Gu, and H.~Hu.
\newblock Automatic generation of synthetic lidar point clouds for 3-d data
  analysis.
\newblock {\em IEEE Transactions on Instrumentation and Measurement}, 2019.

\bibitem{wong2019identifying}
K.~Wong, S.~Wang, M.~Ren, M.~Liang, and R.~Urtasun.
\newblock Identifying unknown instances for autonomous driving.
\newblock {\em arXiv}, 2019.

\bibitem{wymann2000torcs}
B.~Wymann, E.~Espi{\'e}, C.~Guionneau, C.~Dimitrakakis, R.~Coulom, and
  A.~Sumner.
\newblock Torcs, the open racing car simulator.
\newblock {\em Software available at http://torcs. sourceforge. net}, 2000.

\bibitem{xia2019interactive}
F.~Xia, W.~B. Shen, C.~Li, P.~Kasimbeg, M.~Tchapmi, A.~Toshev,
  R.~Mart{\'\i}n-Mart{\'\i}n, and S.~Savarese.
\newblock Interactive gibson: A benchmark for interactive navigation in
  cluttered environments.
\newblock {\em arXiv}, 2019.

\bibitem{xia2018gibson}
F.~Xia, A.~R. Zamir, Z.~He, A.~Sax, J.~Malik, and S.~Savarese.
\newblock Gibson env: Real-world perception for embodied agents.
\newblock In {\em CVPR}, 2018.

\bibitem{yang2018hdnet}
B.~Yang, M.~Liang, and R.~Urtasun.
\newblock Hdnet: Exploiting hd maps for 3d object detection.
\newblock In {\em CoRL}, 2018.

\bibitem{yang2019pointflow}
G.~Yang, X.~Huang, Z.~Hao, M.-Y. Liu, S.~Belongie, and B.~Hariharan.
\newblock Pointflow: 3d point cloud generation with continuous normalizing
  flows.
\newblock {\em arXiv}, 2019.

\bibitem{yoon2019mapless}
D.~Yoon, T.~Tang, and T.~Barfoot.
\newblock Mapless online detection of dynamic objects in 3d lidar.
\newblock In {\em CRV}, 2019.

\bibitem{yue2018lidar}
X.~Yue, B.~Wu, S.~A. Seshia, K.~Keutzer, and A.~L. Sangiovanni-Vincentelli.
\newblock A lidar point cloud generator: from a virtual world to autonomous
  driving.
\newblock In {\em ICMR}, 2018.

\bibitem{zeng2019end}
W.~Zeng, W.~Luo, S.~Suo, A.~Sadat, B.~Yang, S.~Casas, and R.~Urtasun.
\newblock End-to-end interpretable neural motion planner.
\newblock In {\em CVPR}, 2019.

\bibitem{chrisvoxel}
C.~Zhang, W.~Luo, and R.~Urtasun.
\newblock Efficient convolutions for real-time semantic segmentation of 3d
  point clouds.
\newblock In {\em 3DV}, 2018.

\bibitem{zhu2017target}
Y.~Zhu, R.~Mottaghi, E.~Kolve, J.~J. Lim, A.~Gupta, L.~Fei-Fei, and A.~Farhadi.
\newblock Target-driven visual navigation in indoor scenes using deep
  reinforcement learning.
\newblock In {\em ICRA}, 2017.

\end{thebibliography}
}

\end{document}